\documentclass[conference]{IEEEtran}
\IEEEoverridecommandlockouts
\usepackage{cite}
\usepackage{amsmath,amssymb,amsfonts}
\usepackage{algorithmic}
\usepackage{float}
\usepackage{caption}
\usepackage{subcaption}
\usepackage{graphicx}
\usepackage{textcomp}
\usepackage{hyperref}
\usepackage{xcolor}
\usepackage{url}
\usepackage{booktabs}
\def\BibTeX{{\rm B\kern-.05em{\sc i\kern-.025em b}\kern-.08em
    T\kern-.1667em\lower.7ex\hbox{E}\kern-.125emX}}
\begin{document}

\title{%
  \resizebox{1\linewidth}{!}{SBAMP: Sampling Based Adaptive Motion Planning}%
\thanks{
The authors are with the General Robotics, Automation, Sensing and
Perception (GRASP) Laboratory, University of Pennsylvania, Philadelphia, PA, 19104, USA.}
\thanks{
${^*}$ Equal contribution.}
\thanks{
${^\dagger}$ Corresponding author: \href{mailto:quanpham@seas.upenn.edu}{quanpham@seas.upenn.edu}.}
\thanks{Code \& videos available at: \url{https://github.com/anhquanpham/SBAMP}.}
}

\author{
\IEEEauthorblockN{Shreyas Raorane${^*}$, Kabir Ram Puri${^*}$, Anh-Quan Pham${^\dagger}$}
}

\maketitle

\begin{abstract}
Autonomous robots operating in dynamic environments must balance global path optimality with real-time responsiveness to disturbances. This requires addressing a fundamental trade-off between computationally expensive global planning and fast local adaptation. Sampling-based planners such as RRT* produce near-optimal paths but struggle under perturbations, while dynamical systems approaches like SEDS enable smooth reactive behavior but rely on offline data-driven optimization. We introduce Sampling-Based Adaptive Motion Planning (SBAMP), a hybrid framework that combines RRT*-based global planning with an online, Lyapunov-stable SEDS-inspired controller that requires no pre-trained data. By integrating lightweight constrained optimization into the control loop, SBAMP enables stable, real-time adaptation while preserving global path structure. Experiments in simulation and on RoboRacer hardware demonstrate robust recovery from disturbances, reliable obstacle handling, and consistent performance under dynamic conditions.
\end{abstract}


\section{Introduction}
Autonomous robots must navigate geometrically complex and dynamically changing environments, including dodging pedestrians, rerouting around debris, and recovering from sudden collisions, demanding two competing capabilities: \emph{global path quality} (near-optimal, collision-free trajectories over long horizons) and \emph{local reactivity} (instantaneous adaptation to new obstacles or perturbations).

Sampling-based planners like RRT*~\cite{karaman2011sampling} guarantee asymptotic optimality in static scenes but incur significant overhead when replanning under change. Reactive controllers such as SEDS~\cite{khansari2011learning} and LPV-DS~\cite{figueroa2018physically} offer smooth, real-time adaptation but rely on offline demonstrations.

We present \emph{Sampling-Based Adaptive Motion Planning} (SBAMP), a hybrid framework that fits a Lyapunov-stable vector field online to each RRT* waypoint segment, requiring no pre-collected data, and interleaves high-rate local control with lower-frequency global replanning to avoid expensive full replanning. Our main contributions are:
\begin{itemize}
  \item \emph{A bi-level SBAMP architecture} combining RRT* global planning with an online, Lyapunov-stable SEDS-inspired controller that rescues RRT* under severe perturbations.
  \item \emph{An efficient interleaving scheme} minimizing global replanning while preserving provable local stability.
  \item \emph{Extensive evaluation on RoboRacer~\cite{okelly2020f1tenth}} hardware and simulation, showcasing rapid disturbance recovery and robust obstacle resilience.
\end{itemize}

\section{Related Work}

Sampling-Based Motion Planning (SBMP) algorithms such as RRT* enable efficient, collision-free path planning in high-dimensional spaces and converge asymptotically to optimal solutions~\cite{lavalle2006planning,karaman2011anytime,arxiv2023sampling}. Variants including bi-directional RRT* and heuristic-enhanced methods further accelerate convergence~\cite{akgun2011sampling}. However, classical SBMP methods are inherently static: once a path is generated, they lack mechanisms for real-time adaptation to environmental changes, and extensions handling kinodynamic or dynamic constraints often sacrifice online practicality.

Learning-based Dynamical Systems (DS) address adaptability by modeling robot motion as stable attractor systems. SEDS fits a Gaussian mixture model to demonstration data under Lyapunov constraints, ensuring global asymptotic stability~\cite{khansari2011learning}, while LPV-DS generalizes this via state-dependent linear models with stability certificates across operating regimes~\cite{figueroa2018physically}. GP-MDS enables online refinement through Gaussian Process Regression without batch training, though it requires careful kernel tuning and sparse-data management~\cite{kronander2015incremental}. All three paradigms share a core limitation: dependence on offline demonstrations, extensive dataset collection, or nontrivial model tuning, which hinders integration with global planners in real-time, unstructured environments.

Hybrid frameworks attempt to combine global exploration with local adaptability. Recent work couples RRT* with Lyapunov-certified, demonstration-driven controllers to funnel around nominal waypoints~\cite{arxiv2023sampling}, but relies on pre-collected data and lacks a unified global stability guarantee. Robust samplers incorporating forward reachability analysis~\cite{wu2022robustrrt} similarly omit Lyapunov-style proofs, while chance-constrained RRT variants using tube-based LPV-MPC~\cite{nezami2022robust} and LPV-embedded nonlinear MPC~\cite{karachalios2025efficient} achieve probabilistic robustness but incur significant per-step optimization overhead. None of these methods simultaneously eliminate demonstration dependence, guarantee unified stability, and maintain real-time tractability.

SBAMP addresses all three gaps by fitting its SEDS-style Gaussian mixture model on-the-fly from each newly planned RRT* segment—requiring no offline data—and synthesizing every local controller under a common Lyapunov-style local stability constraint. By decoupling RRT* planning from vector-field evaluation (a weighted sum of linear maps at control rate), SBAMP sustains real-time performance without per-step optimization solves.

\section{Sampling Based Adaptive Motion Planning}
\label{sec:SBAMP-theory}

\begin{figure}[H]
\centering
\includegraphics[width=1\columnwidth]{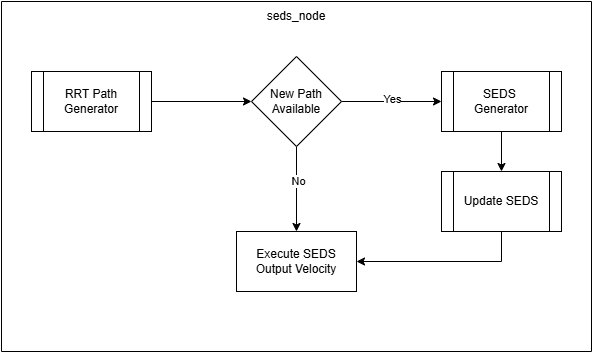}
\caption{Flowchart of the SBAMP theoretical framework. When \emph{New Path Available?} is true, the SEDS generator refits the dynamical system to the latest RRT$^*$ segment; otherwise, the existing SEDS velocity command is executed.}
\label{fig:SBAMP-theory}
\end{figure}

Figure~\ref{fig:SBAMP-theory} depicts the overall SBAMP control loop. At its core, SBAMP runs two modules in parallel: a global RRT$^*$ planner and a local SEDS controller, with a lightweight decision logic that refits the dynamical system whenever the planner produces a new path.

SBAMP is structured as a bi-level optimization framework comprising three interacting components.

\subsection{Global Path Planning via RRT$^*$}
We incrementally grow and rewire a tree $\mathcal{T}\subset\mathcal{C}_{\rm free}$ using a lightweight RRT$^*$-inspired planner~\cite{karaman2011anytime}, yielding a waypoint sequence
\[
  \tau = \{\,x_0,\,x_1,\,\dots,\,x_g\}\subset\mathbb R^n.
\]
Every planner cycle (period $\Delta t_G$) samples $x_{\rm rand}\sim\mathcal{U}(\mathcal{C}_{\rm free})$, extends toward it, and performs local rewiring over nearby nodes to improve path optimality.

\subsection{Local Trajectory Adaptation via SEDS}
At control rate ($\Delta t_C \ll \Delta t_G$), the robot state $\xi(t)$ is driven by a convex mixture of $K$ linear subsystems~\cite{khansari2011learning}:
\begin{equation}\label{eq:seds-theory}
  \dot{\xi}
  = f(\xi)
  = \sum_{k=1}^K \gamma_k(\xi)\bigl(A_k\,\xi + b_k\bigr),
  \quad
  \sum_{k=1}^K\gamma_k(\xi)=1,\;\gamma_k(\xi)\ge0,
\end{equation}
where each $A_k\!+\!A_k^\top\prec0$ (ensuring $V(\xi)=\xi^\top\xi$ decays) and
\[
  b_k = -A_k\,x_i \quad\Longrightarrow\quad f(x_i)=0
\]
at the active waypoint $x_i$. We pose local controller synthesis as a real-time constrained optimization problem: given waypoints $\{x_i, x_{i+1}\}$, find a $K$-component mixture of linear systems that (1) produces smooth vector fields via GMM fitting, and (2) guarantees local asymptotic stability via Hurwitz projection. Formally, for each component $k$:
\[
  \min_{A_k}\;\|A_k - \hat{A}_k\|_F \quad \text{s.t.} \quad A_k + A_k^\top \prec 0,
\]
This per-cycle fit requires no stored dataset and completes within the replanning loop. The attractor is then recentered at $x_{i+1}$ by updating $\{b_k\}$ accordingly.

\subsection{Real-Time Integration and Stability}
Upon receiving a new path, the SEDS generator refits $\{b_k\}$ and updates $\dot{\xi}$; otherwise, the current model is used, with attractor shifts preserving velocity continuity.
Under the average dwell-time theorem~\cite{hespanha1999stability}, if the SEDS update period $\Delta t_C$ and RRT$^*$ planning period $\Delta t_G$ satisfy
\[
  \Delta t_C \;\ll\;\tau_D\;\le\;\Delta t_G,
\]
then each fitted subsystem remains locally stable to its active waypoint, with empirically observed recovery to the final goal $x_g$. Together, these three modules realize a real-time adaptive planner with Lyapunov-stable local subsystems and no offline training data required. Full implementation details on the RoboRacer \cite{okelly2020f1tenth} hardware are provided in Appendix~\ref{sec:appendix-impl}.

\section{Experiments}

We evaluate SBAMP against standard RRT* across three complementary studies designed to stress-test the two core claims of the framework: that the DS attractor preserves control authority during replanning gaps, and that Lyapunov-stable local control enables recovery from disturbances that exceed RRT*'s planning capacity.

\subsection{Computational Efficiency Under Disturbance}
\label{sec:exp1}

A fundamental vulnerability of purely sampling-based planners is that replanning latency grows with perturbation magnitude: as the vehicle is displaced farther from its prior tree, RRT* must explore a larger region of $\mathcal{C}_{\rm free}$ before recovering a feasible path, and if that latency exceeds the time to exhaust the current waypoint buffer, the controller loses its reference entirely. We quantify this degradation by teleporting the vehicle laterally by $\Delta d \in [2.25, 2.75]$\,m at a fixed straightaway immediately before each planning cycle, repeating $N=20$ trials per displacement in the ROS2 simulator at $v = 1$\,m/s.

\begin{figure}[H]
  \centering
  \includegraphics[width=0.75\columnwidth]{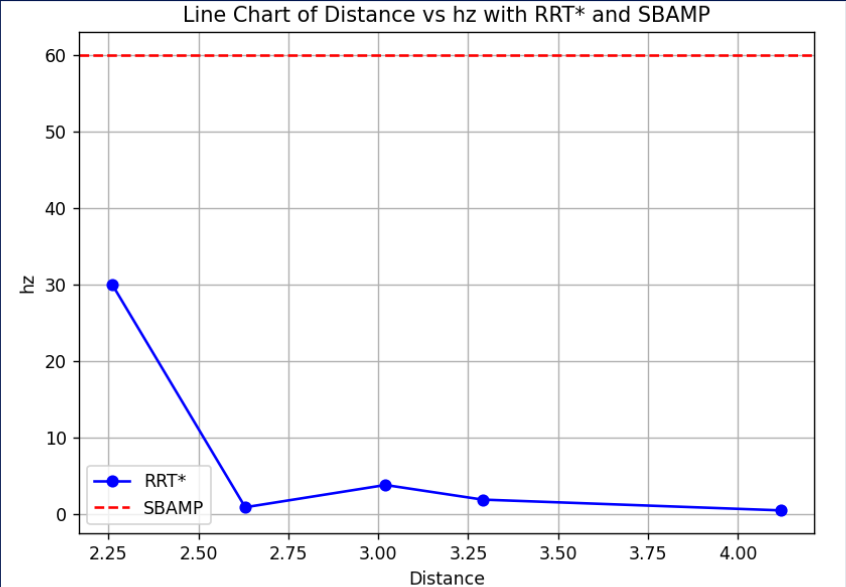}
  \caption{Replanning frequency vs. lateral perturbation. RRT* falls below the 2\,Hz stability threshold as $\Delta d$ grows; SBAMP maintains approximately 60\,Hz throughout.}
  \label{fig:replan_freq_vs_perturbation}
\end{figure}

Figure~\ref{fig:replan_freq_vs_perturbation} reports replanning frequency $f_{\rm plan}$ as a function of $\Delta d$. RRT* degrades sharply across this range, falling below the 2\,Hz minimum required to ensure the vehicle advances no more than 0.5\,m between updates at nominal speed. SBAMP, by contrast, sustains approximately 60\,Hz throughout, because the DS attractor provides a well-defined velocity command toward the last known waypoint irrespective of planner latency, and transitions to each new RRT* solution without discontinuity in the control signal. This 30$\times$ margin over the stability threshold directly validates the dwell-time argument of Section~\ref{sec:SBAMP-theory}.

\subsection{Recovery from Extreme Planner Failures}
\label{sec:exp2}

To characterize the boundary of RRT*'s recovery envelope and demonstrate SBAMP's behavior beyond it, we subjected both planners to three qualitatively distinct failure modes in a $5\,\text{m} \times 2\,\text{m}$ corridor: large translational jumps, rotational offsets up to $90^\circ$, and corner entrapment. In each case we increased disturbance magnitude until the planner either collided or failed to produce a feasible path within the planning budget.

\begin{figure}[H]
  \centering
  \begin{subfigure}[t]{0.48\linewidth}
    \centering
    \includegraphics[width=\linewidth]{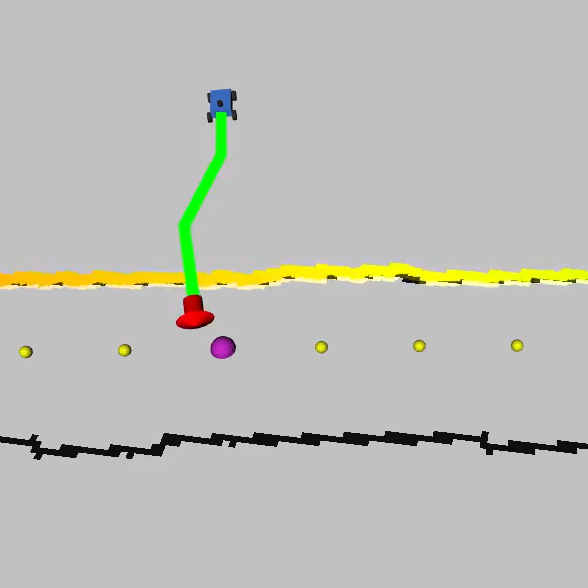}
    \caption{RRT* under large translation}
    \label{fig:RRT_large_trans}
  \end{subfigure}%
  \hfill
  \begin{subfigure}[t]{0.48\linewidth}
    \centering
    \includegraphics[width=\linewidth]{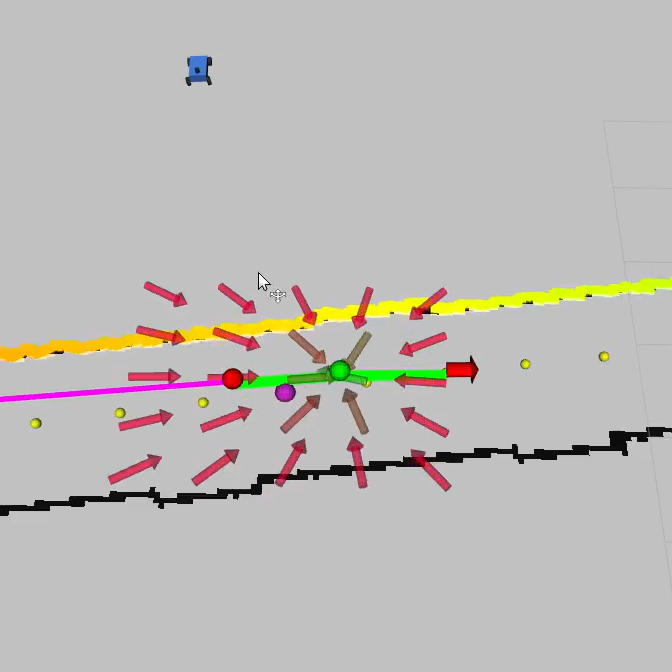}
    \caption{SBAMP under large translation}
    \label{fig:SBAMP_large_trans}
  \end{subfigure}
  \caption{Planner recovery under large translational perturbation. RRT* loses its reference; SBAMP maintains a stable attractor throughout.}
  \label{fig:pair_large_trans}
\end{figure}

Under large lateral displacement (Figure~\ref{fig:pair_large_trans}), RRT* exhausts its planning budget before reconnecting to the corridor, leaving the vehicle with no valid reference. SBAMP's attractor immediately redirects the vehicle toward the last RRT* waypoint, maintaining bounded tracking error until the planner recovers and a new path is handed off without discontinuity.

\begin{figure}[H]
  \centering
  \begin{subfigure}[t]{0.48\linewidth}
    \centering
    \includegraphics[width=\linewidth]{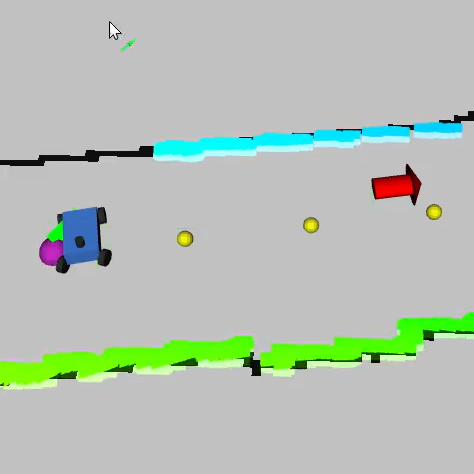}
    \caption{RRT* under large rotation}
    \label{fig:RRT_large_rotate}
  \end{subfigure}%
  \hfill
  \begin{subfigure}[t]{0.48\linewidth}
    \centering
    \includegraphics[width=\linewidth]{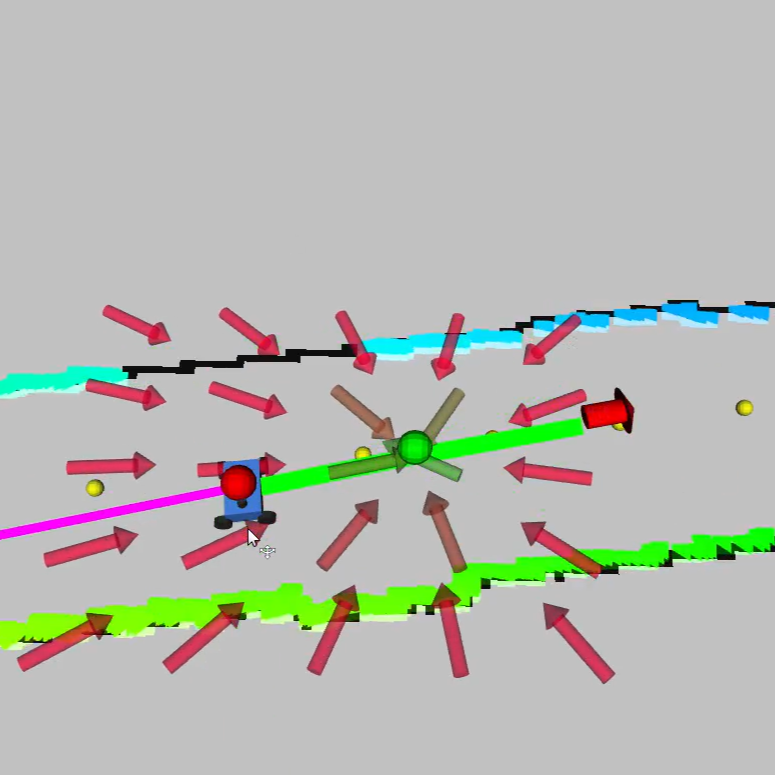}
    \caption{SBAMP under large rotation}
    \label{fig:SBAMP_large_rotate}
  \end{subfigure}
  \caption{Planner recovery under large rotational perturbation. RRT* produces unsafe paths; SBAMP maintains stable convergence to the last waypoint.}
  \label{fig:pair_large_rotate}
\end{figure}

Rotational offsets exceeding $60^\circ$ expose a second failure mode: RRT* either times out or produces paths that, when executed, direct the vehicle into obstacles before a corrective replan can arrive (Figure~\ref{fig:pair_large_rotate}). SBAMP is unaffected because the DS controller operates on the error to the current waypoint in Cartesian space, not on heading, and continues issuing stable commands regardless of orientation.

\begin{figure}[H]
  \centering
  \begin{subfigure}[t]{0.48\linewidth}
    \centering
    \includegraphics[width=\linewidth]{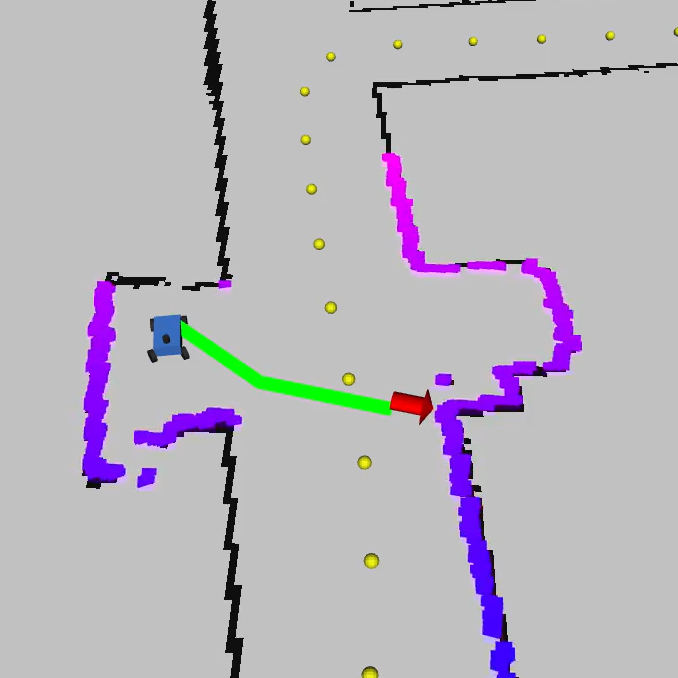}
    \caption{RRT* in tight corners}
    \label{fig:RRT_tight_corners}
  \end{subfigure}%
  \hfill
  \begin{subfigure}[t]{0.48\linewidth}
    \centering
    \includegraphics[width=\linewidth]{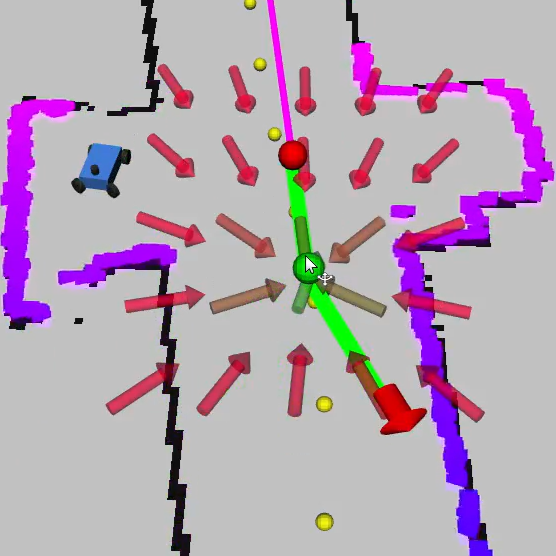}
    \caption{SBAMP in tight corners}
    \label{fig:SBAMP_tight_corners}
  \end{subfigure}
  \caption{Performance in tight-corner scenarios. RRT* produces unsafe waypoints; SBAMP commits only to the immediately reachable target.}
  \label{fig:pair_tight_corners}
\end{figure}

In tight-corner scenarios (Figure~\ref{fig:pair_tight_corners}), RRT*'s sparse sampling produces waypoints that, under execution, bring the vehicle within collision range of opposing walls. SBAMP resolves this by committing only to the immediately reachable waypoint via the SEDS vector field and withholding progression until the next global plan is available, yielding smooth, collision-free negotiation of the corner.

\begin{figure}[H]
  \centering
  \begin{subfigure}[t]{0.48\linewidth}
    \centering
    \includegraphics[width=\linewidth]{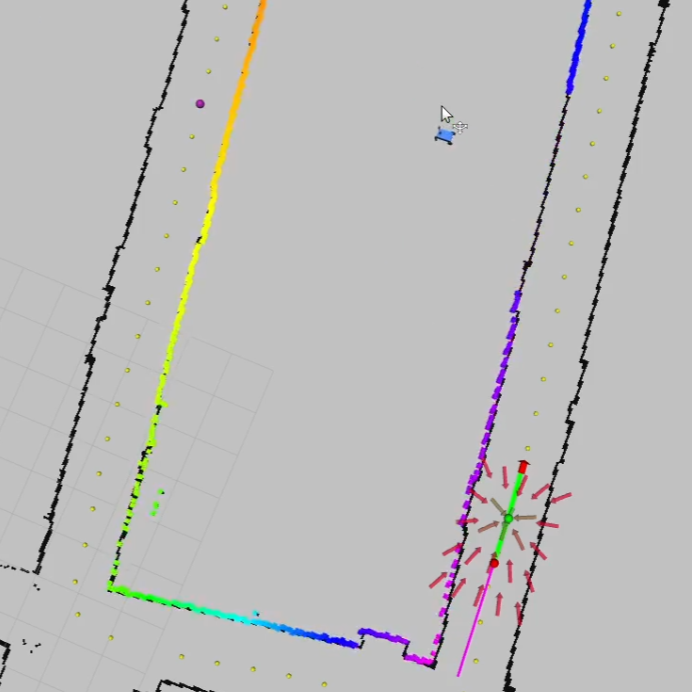}
    \caption{Pre-recovery state}
    \label{fig:SBAMP_recovery_before}
  \end{subfigure}%
  \hfill
  \begin{subfigure}[t]{0.48\linewidth}
    \centering
    \includegraphics[width=\linewidth]{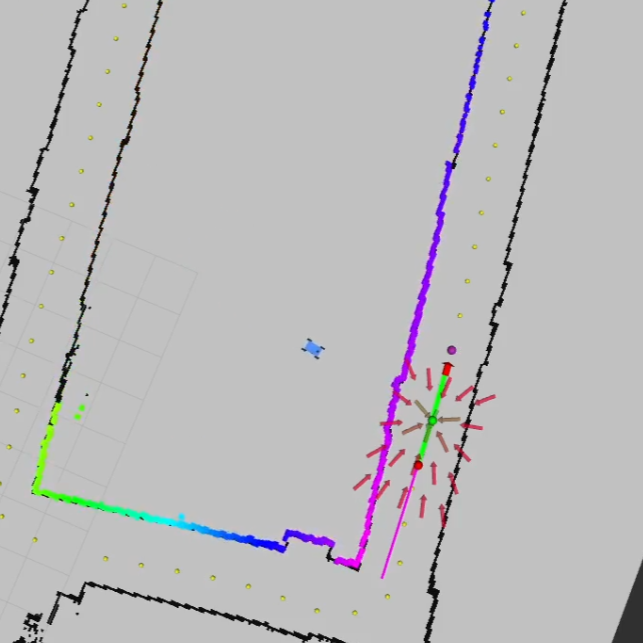}
    \caption{Post-recovery state}
    \label{fig:SBAMP_recovery_after}
  \end{subfigure}
  \caption{SBAMP recovery under combined large translational and rotational displacement.}
  \label{fig:SBAMP_large_recovery}
\end{figure}

SBAMP's recovery is not limited to small perturbations: even under large translational and rotational displacements (Figure~\ref{fig:SBAMP_large_recovery}), the SEDS attractor guides the vehicle back into the connected free-space corridor, at which point a new global trajectory is computed and followed without discontinuity.

\subsection{Real-Time Performance Validation on Hardware}
\label{sec:exp3}

Simulation results establish SBAMP's theoretical properties; hardware trials establish that these properties transfer to a physical platform subject to sensor noise, actuation lag, and unmodeled dynamics. We operated the vehicle on an indoor loop course featuring straightaways, tight turns, and cluttered corridors, and manually applied 20 randomized translational and rotational disturbances during closed-loop operation.

\begin{figure}[H]
  \centering
  \begin{subfigure}[t]{0.48\linewidth}
    \centering
    \includegraphics[width=\linewidth]{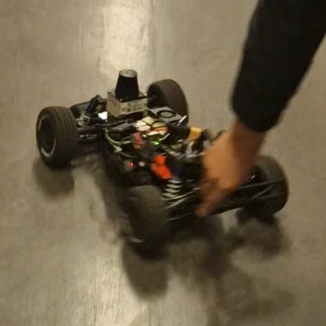}
    \caption{Post-disturbance deviation}
    \label{fig:SBAMP_human_rotate_before}
  \end{subfigure}%
  \hfill
  \begin{subfigure}[t]{0.48\linewidth}
    \centering
    \includegraphics[width=\linewidth]{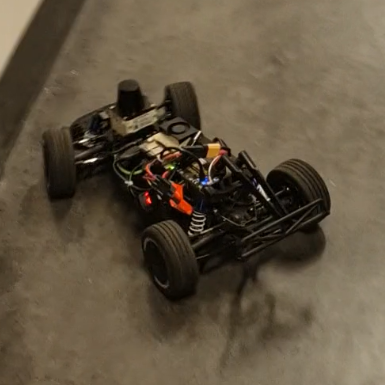}
    \caption{Post-recovery trajectory}
    \label{fig:SBAMP_human_rotate_after}
  \end{subfigure}
  \caption{SBAMP response to human-applied rotational disturbance on hardware.}
  \label{fig:pair_human_rotate}
\end{figure}

\begin{figure}[H]
  \centering
  \begin{subfigure}[t]{0.48\linewidth}
    \centering
    \includegraphics[width=\linewidth]{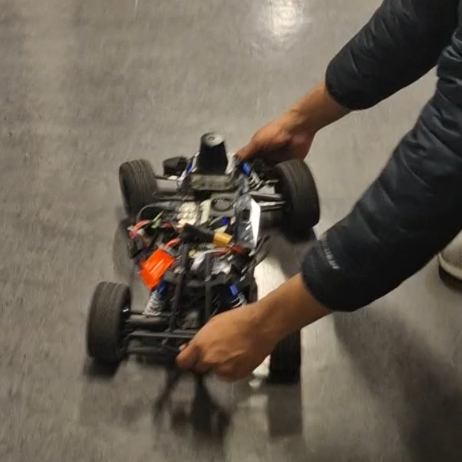}
    \caption{Post-disturbance deviation}
    \label{fig:SBAMP_human_translate_before}
  \end{subfigure}%
  \hfill
  \begin{subfigure}[t]{0.48\linewidth}
    \centering
    \includegraphics[width=\linewidth]{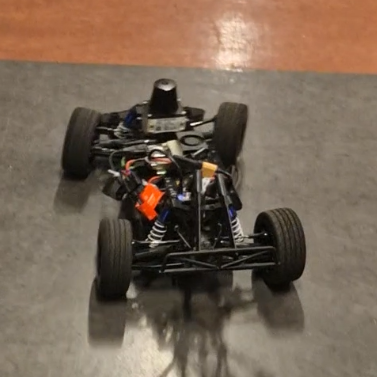}
    \caption{Post-recovery trajectory}
    \label{fig:SBAMP_human_translate_after}
  \end{subfigure}
  \caption{SBAMP response to human-applied translational disturbance on hardware.}
  \label{fig:pair_human_translate}
\end{figure}

In all trials, the vehicle deviated from its nominal path immediately following the disturbance, after which the SEDS attractor generated commands that returned it to the vicinity of the last RRT* waypoint before transitioning seamlessly to the newly computed global plan (Figures~\ref{fig:pair_human_rotate}--\ref{fig:pair_human_translate}). SBAMP achieved a near-100\% recovery rate across all 20 perturbations; the isolated failures arose only under extreme rotational displacements sufficient to cause waypoint misidentification, and even in these cases the vehicle converged to the most recently valid waypoint rather than diverging.

\begin{figure}[H]
  \centering
  \begin{subfigure}[t]{0.48\linewidth}
    \centering
    \includegraphics[width=\linewidth]{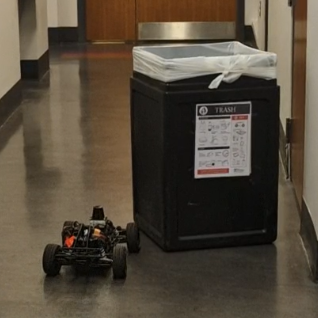}
    \caption{Approaching obstacle (A)}
    \label{fig:SBAMP_obsAvoid_A_before}
  \end{subfigure}%
  \hfill
  \begin{subfigure}[t]{0.48\linewidth}
    \centering
    \includegraphics[width=\linewidth]{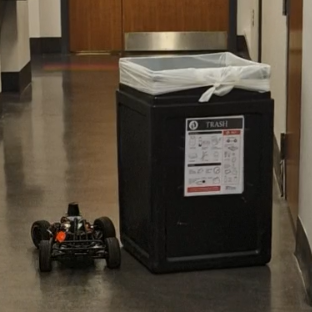}
    \caption{Left-side avoidance (A)}
    \label{fig:SBAMP_obsAvoid_A_after}
  \end{subfigure}
  \caption{Obstacle avoidance scenario A.}
  \label{fig:pair_obsAvoid_A}
\end{figure}

\begin{figure}[H]
  \centering
  \begin{subfigure}[t]{0.48\linewidth}
    \centering
    \includegraphics[width=\linewidth]{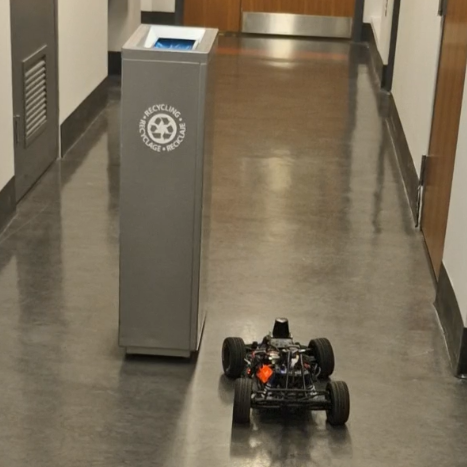}
    \caption{Approaching obstacle (B)}
    \label{fig:SBAMP_obsAvoid_B_before}
  \end{subfigure}%
  \hfill
  \begin{subfigure}[t]{0.48\linewidth}
    \centering
    \includegraphics[width=\linewidth]{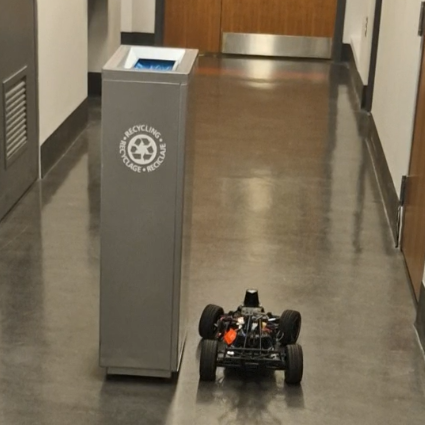}
    \caption{Right-side avoidance (B)}
    \label{fig:SBAMP_obsAvoid_B_after}
  \end{subfigure}
  \caption{Obstacle avoidance scenario B.}
  \label{fig:pair_obsAvoid_B}
\end{figure}

Finally, Figures~\ref{fig:pair_obsAvoid_A}--\ref{fig:pair_obsAvoid_B} demonstrate real-world obstacle avoidance under two drift scenarios. In each case SBAMP generated a collision-free avoidance trajectory around an unexpected object and rejoined the nominal corridor, without any modification to the underlying RRT* planner. This non-invasive augmentation is a key property of the framework: when RRT* operates correctly, SBAMP defers to it; when the planner falters, the DS attractor intervenes autonomously.

\section{Conclusion}

We have introduced SBAMP, a bi-level motion-planning framework that non-invasively augments RRT* with a Lyapunov-stable dynamical-systems controller, achieving on-the-fly adaptation with no prior training data. By converting each RRT* waypoint into a locally stable attractor on-the-fly, SBAMP ensures a valid control reference even when global replanning lags. Our threefold evaluation demonstrates that SBAMP sustains high replanning frequencies, reliably recovers from large translational and rotational disturbances, and executes safe obstacle avoidance, all without any offline learning or demonstration dataset.

Future work includes integrating SBAMP with receding-horizon optimizers such as MPC or MPPI, embedding obstacle-repulsive modulation directly into the dynamical-systems layer to reduce reliance on occupancy-grid update rates, and extending the framework to high-dimensional manipulators to broaden its applicability across autonomous robotics tasks.

\bibliographystyle{IEEEtran}
\bibliography{refs}

\appendices
\section{SBAMP Implementation on RoboRacer}
\label{sec:appendix-impl}

SBAMP is deployed on the RoboRacer~\cite{okelly2020f1tenth} platform using ROS2 Humble. Laser scans and odometry feed into a local occupancy grid; the planner produces a waypoint sequence $\tau$; the SEDS controller issues velocity commands; and an optional visualization node renders the state in RViz2.

Five ROS2 nodes form the backbone of SBAMP: an \emph{Occupancy Grid Node} fusing LIDAR and odometry; a \emph{Next Waypoint Node} extracting the next feasible goal; an \emph{RRT* Node} continuously replanning collision-free paths; an optional \emph{Visualization Node} for RViz2 debugging; and the \emph{SBAMP Node} fitting the SEDS model and publishing Ackermann drive commands at high frequency. The five core nodes are listed in Table~\ref{tab:ros2-packages}.

\begin{figure}[H]
  \centering
  \includegraphics[width=0.9\columnwidth]{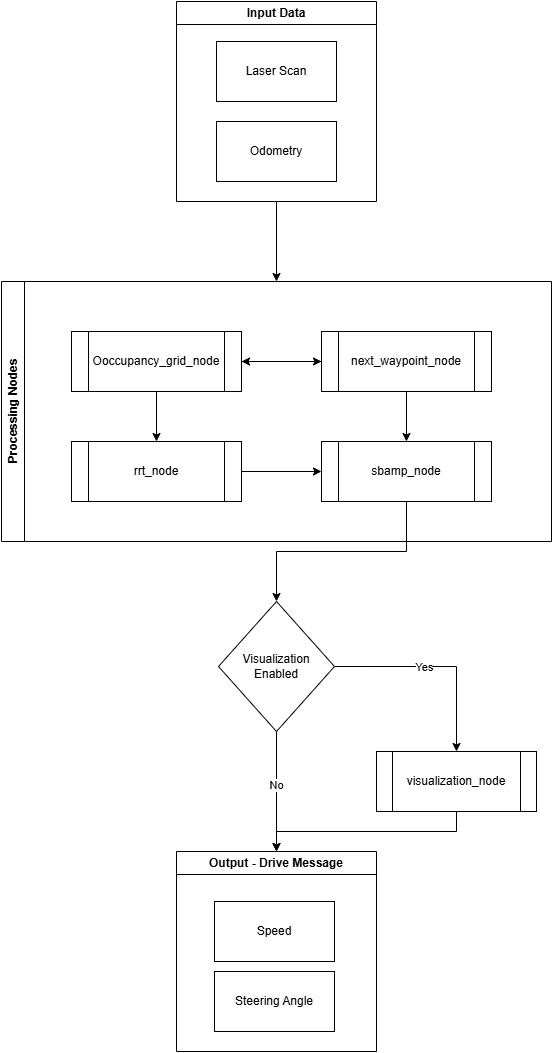}
  \caption{ROS2 node graph for SBAMP on RoboRacer.}
  \label{fig:SBAMP-ROS-RoboRacer}
\end{figure}

\begin{table}[ht]
  \centering
  \caption{Core ROS2 Nodes in the \texttt{sbamp} Package}
  \label{tab:ros2-packages}
  \begin{tabular}{@{}ll@{}}
    \toprule
    \textbf{Node}                    & \textbf{Functionality}                  \\
    \midrule
    \texttt{occupancy\_grid\_node}   & LIDAR/odometry fusion                   \\
    \texttt{rrt\_node}               & Online path planning                    \\
    \texttt{next\_waypoint\_node}    & Feasible goal extraction                \\
    \texttt{sbamp\_node}             & SEDS fit and Ackermann control          \\
    \texttt{visualization\_node}     & RViz2 rendering (optional)              \\
    \bottomrule
  \end{tabular}
\end{table}

All experiments were performed on the RoboRacer F1/10 platform, whose kinematics obey
\[
\dot{x} = v\cos\theta,\quad
\dot{y} = v\sin\theta,\quad
\dot{\theta} = \frac{v}{L}\tan\delta.
\]
Perception is provided by an 812-beam SICK TIM781 LIDAR, and actuation uses ROS2 Ackermann steering commands. 


\end{document}